\documentclass[10pt,twocolumn,letterpaper]{article}

\usepackage[pagenumbers]{cvpr}   
\usepackage{amsmath,amsfonts,amssymb}
\usepackage{mathtools}
\usepackage{graphicx}
\usepackage{array}
\usepackage{booktabs}
\usepackage{multirow}

\usepackage{xcolor}
\usepackage{pifont}
\providecommand{\cmark}{\ding{51}}
\providecommand{\xmark}{\ding{55}}
\providecolor{ourgreen}{RGB}{0,128,0}
\providecolor{ourred}{RGB}{200,0,0}

\usepackage{tabularx}
\usepackage{colortbl}
\definecolor{obrankone}{HTML}{EEF4FA}   
\definecolor{obranktwo}{HTML}{FDF5E9}   
\definecolor{obrankthree}{HTML}{FDFCE2}

\newcommand{\bestcell}[1]{\cellcolor{obrankone}#1}
\newcommand{\secondcell}[1]{\cellcolor{obranktwo}#1}
\newcommand{\thirdcell}[1]{\cellcolor{obrankthree}#1}
\definecolor{cvprblue}{rgb}{0.21,0.49,0.74}
\usepackage[pagebackref,breaklinks,colorlinks,allcolors=cvprblue]{hyperref}

\def\confName{CVPR}
\def\confYear{2027}
\vspace{-3cm}
\title{ObsBridge: Observation-Grounded Conditioning for Ego Trajectory Prediction}
\vspace{-3cm}
\author{%
  Zhenhao Wang\textsuperscript{$\ast$,1},
  Shaofeng Liang\textsuperscript{$\ast$,$\ddagger$,1,2},
  Jiemin Wu\textsuperscript{$\ast$,1},
  Sijia Chen\textsuperscript{3},
  Keni Shi\textsuperscript{1},
  Nan Li\textsuperscript{1},\\
  Wenshuo Chen\textsuperscript{1},
  Bowen Tian\textsuperscript{1,2},
  Runwei Guan\textsuperscript{1},
  Yutao Yue\textsuperscript{1,4,$\dagger$}\\
  {\small\textsuperscript{1}The Hong Kong University of Science and Technology (Guangzhou), \small\textsuperscript{2}EigenAct Technology Co., Ltd.}\\
  {\small\textsuperscript{3}Huazhong University of Science and Technology\quad
   \textsuperscript{4}Institute of Deep Perception Technology (JITRI)}\\
  {\small\textsuperscript{$\ast$}Equal contribution\quad
   \textsuperscript{$\ddagger$}Project lead\quad
   \textsuperscript{$\dagger$}Corresponding author}
}

\begin{document}
\twocolumn[{
\maketitle
\begin{center}
    \captionsetup{type=figure}
    \includegraphics[width=1.0\textwidth]{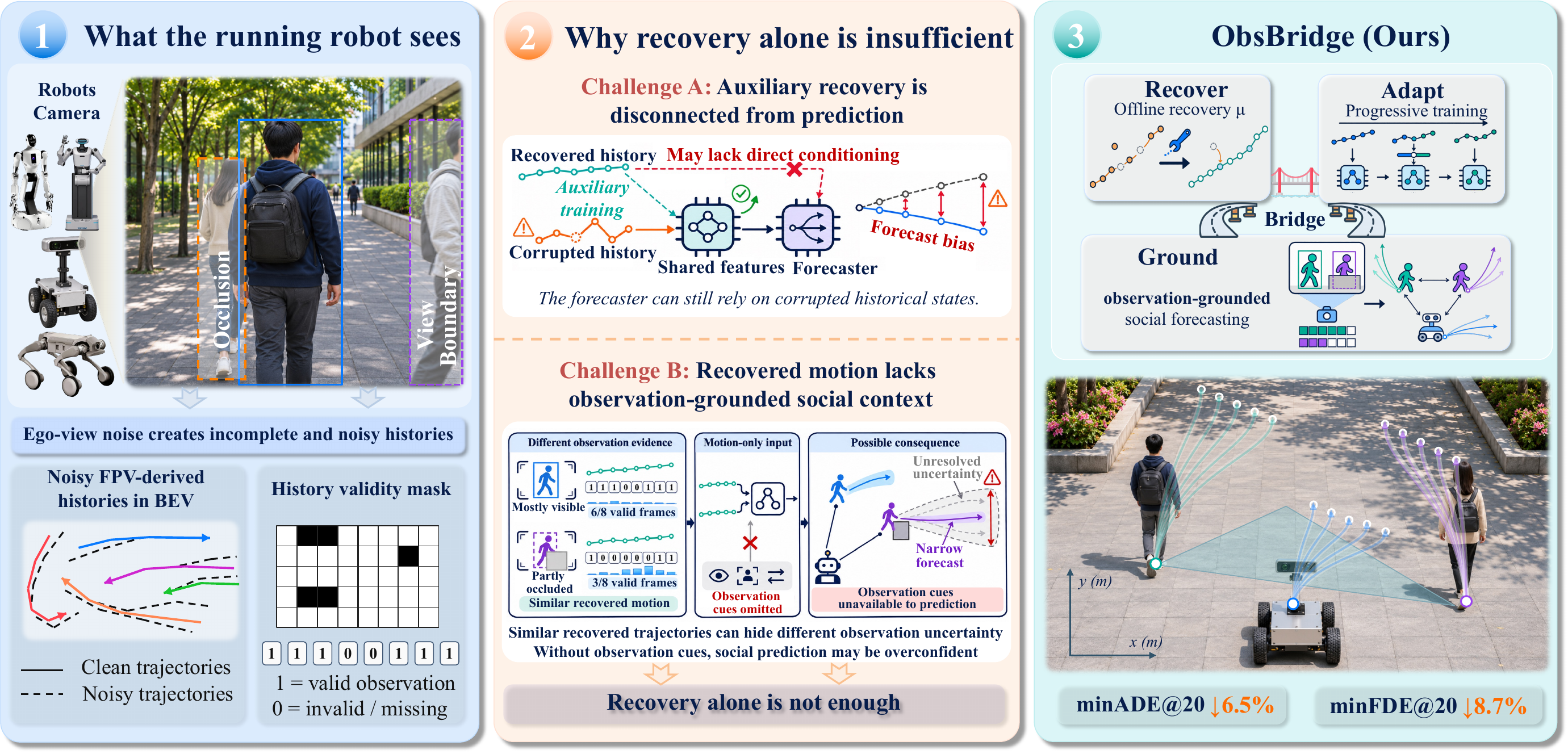}
   \caption{\textbf{Motivation and overview of ObsBridge.} \textbf{(1)} Ego-view perception yields incomplete and noisy historical trajectories in BEV. \textbf{(2)} Auxiliary reconstruction lacks direct conditioning for prediction (Challenge A), while pure coordinate recovery omits observation conditions (Challenge B). \textbf{(3)} \textbf{ObsBridge} couples offline recovery, progressive curriculum adaptation, and observation-grounded social context to substantially reduce forecasting errors.}
   \label{fig:teaser}
\end{center}
}]
\maketitle
\begin{abstract}
Pedestrian forecasting from ego-view observations is challenging because occlusion, limited visibility, and tracking errors corrupt observed motion histories. Such errors are particularly consequential for autonomous robot navigation. Existing paradigms typically either impute noisy history positions before forecasting or jointly train an auxiliary history reconstruction branch. 
However, predictors trained on clean histories may be sensitive to residual errors in recovered histories at deployment.
Meanwhile, auxiliary reconstruction still computes inter-agent interactions over raw, corrupted coordinates. 
We investigate conditioning future prediction on both recovered historical states and raw observation evidence, with progressive training to accommodate imperfect recovery.
Building on this insight, we propose ObsBridge, a flow-based trajectory forecasting framework with observation-grounded historical bridging. Specifically, ObsBridge first employs a frozen estimator to recover historical motion from agent-relative and ego-relative positions, image-plane locations, and visibility states. During training, we introduce an agent-wise curriculum that gradually replaces clean teacher histories with complete estimated histories before context encoding. 
This progressively exposes the predictor to estimated histories during training.
Furthermore, we design a residual geometry adapter to fuse the encoded motion with original observation evidence. We aggregate the resulting representations through social attention to condition the flow-based trajectory generator.
Extensive experiments across multiple benchmarks demonstrate that ObsBridge outperforms state-of-the-art methods.
On EgoTraj-TBD, it reduces minADE@20 by 6.5\% and minFDE@20 by 8.7\%.  The code will be provided at 
\href{https://github.com/shawnliang420/ObsBridge}{https://github.com/shawnliang420/ObsBridge}.
\end{abstract}    
\vspace{-2em}
\section{Introduction}
Forecasting the future trajectories of surrounding pedestrians is a fundamental capability for autonomous systems, such as mobile service robots, operating in dynamic human environments \cite{alahi2016sociallstm, gupta2018socialgan, liang2025cognitive, poddar2023crowd}. Most existing trajectory predictors achieve strong accuracy under idealized bird-eye-view (BEV) settings with complete and noiseless observations \cite{stoler2023t2fpv, liu2026egotraj}. However, these idealized assumptions rarely hold in real-world deployment.
In practice, onboard perception relies on first-person-view (FPV) cameras. Historical trajectories extracted from these camera observations are frequently corrupted by mutual occlusions, field-of-view truncations, tracking switches, and perspective distortion.
When these trajectories are projected into global metric coordinates, the perceptual artifacts introduce substantial errors in historical positions. These errors severely degrade the performance of standard BEV predictors \cite{liu2026egotraj}.

Recent studies explore two primary directions to handle ego-view observation noise.
The first paradigm employs imputation modules, such as CoFE~\cite{stoler2023t2fpv}, to correct noisy positions before prediction.
When predictors are trained on clean histories, imperfect recovered inputs can introduce a conditioning mismatch at deployment.
Moreover, relying solely on recovered coordinates omits observation reliability, as similar recovered trajectories can stem from different visibility patterns and correction magnitudes.
The second paradigm, represented by BiFlow~\cite{liu2026egotraj}, jointly trains an auxiliary history reconstruction branch alongside future prediction on a shared encoder.
Although this auxiliary objective improves shared features, the reconstructed histories do not condition the predictor at inference.
Consequently, the encoder still computes motion and social representations from corrupted coordinates, leaving the forecaster vulnerable to observation artifacts.

Overcoming these limitations raises two core questions: how to condition future prediction on recovered motion while accounting for residual errors, and how to ground social aggregation in original observation evidence.
Our key insight is to treat recovered histories as useful but imperfect conditioning signals.
Specifically, we progressively adapt the predictor to recovered histories during training, and enrich their encoded motion with ego-relative geometry, image-plane locations, visibility states, and correction cues.
Together, these mechanisms connect historical recovery to observation-grounded social forecasting.

Building on this insight, we propose ObsBridge, a generative trajectory forecasting framework that connects historical recovery to future prediction through observation-grounded bridging. First, a frozen historical estimator recovers complete trajectory corrections from agent-relative and ego-relative positions, normalized image-plane coordinates, and visibility states. 
To bridge the discrepancy between clean training data and imperfect test estimates, we introduce an agent-wise conditional curriculum. During early training, it gradually transitions the context encoder input from clean teacher histories to complete estimated trajectories.
This progression allows the model to establish accurate motion dynamics before smoothly adapting to out-of-fold residual recovery errors. Second, to prevent pure coordinate recovery from obscuring observation reliability, we design a residual geometry adapter followed by a dedicated social attention layer. 
The adapter fuses the encoded motion with per-frame descriptors that encapsulate observed and recovered ego-relative geometry, viewing coordinates, visibility flags, and correction magnitudes. This makes observation evidence explicit within each node representation.
The subsequent social attention layer aggregates these observation-grounded features across agents to condition the dual-stream flow matching generator.
We evaluate ObsBridge across multiple benchmarks, including the real-world EgoTraj-TBD dataset~\cite{liu2026egotraj} and the T2FPV-ETH benchmark~\cite{stoler2023t2fpv}. ObsBridge consistently outperforms state-of-the-art approaches. On EgoTraj-TBD, it reduces minADE@20 by 6.5\% and minFDE@20 by 8.7\%, respectively, while achieving average reductions of 6.3\% and 7.3\% across T2FPV-ETH.

The main contributions are summarized as follows:
\begin{itemize}
    \item We propose ObsBridge, an observation-grounded forecasting framework that bridges historical motion recovery and flow-based prediction under ego-view noise.
    \item We introduce an agent-wise curriculum that transitions historical conditions from clean ground truth to cross-fitted estimates during training.
    \item We introduce a residual geometry adapter and social attention to fuse encoded motion with visibility states, viewing geometry, and correction magnitudes.
\end{itemize}

\section{Related Work}
\label{sec:related_work}

\subsection{Pedestrian Trajectory Prediction}
\label{sec:rw_prediction}
Pedestrian trajectory prediction models multimodal future motion conditioned on observed histories.
Beyond early GAN and CVAE formulations~\cite{gupta2018socialgan,salzmann2020trajectron,sadeghian2019sophie,zhang2024s}, recent diffusion and flow-based predictors advance multimodal representation and sampling efficiency through iterative denoising~\cite{gu2022mid}, leapfrog initialization~\cite{mao2023led}, adaptive anchor refinement in a unified embedding space~\cite{bae2024singulartrajectory}, and one-step distilled flow matching~\cite{fu2025moflow}.
In parallel, social interaction modeling has evolved from early pooling mechanisms~\cite{alahi2016sociallstm,mohamed2020social,bae2022learning} to angular representations~\cite{wong2024socialcircle}, higher-order relational reasoning~\cite{kim2024highgraph}, multi-order intention fusion~\cite{chen2025socialmoif}, and spectral co-vibration decomposition~\cite{wong2025resonance}.
While effective under clean observations, these methods degrade when ego-view histories are missing or corrupted. In contrast, our approach grounds social aggregation in both recovered motion states and raw observation evidence.

\subsection{Trajectory Prediction under Ego-centric Noise}
\label{sec:rw_egocentric}
Ego-centric trajectory forecasting operates either in the image plane using apparent scale and camera motion~\cite{rasouli2024egocentric} or in a shared metric world frame.
To study perception degradation, recent works introduce synthetic first-person benchmarks~\cite{stoler2023t2fpv}, real-world ego-view datasets~\cite{liu2026egotraj}, and perception-aware interfaces incorporating tracking affinities~\cite{weng2022affinipred} or synthetic-to-real adaptation~\cite{fujii2026realtraj}.
To handle incomplete or noisy histories, imputation-based methods recover missing coordinates prior to or jointly with forecasting~\cite{stoler2023t2fpv,chib2024mstip,chib2024trajimpute,chen2026aerialmind}, whereas denoising and reconstruction approaches leverage auxiliary objectives~\cite{li2025natra,qin2026nim} or dual-stream flow matching~\cite{liu2026egotraj}.
However, coordinate-only imputation does not explicitly capture observation reliability, while auxiliary reconstruction branches fail to route denoised trajectories back to condition the future predictor.
Our framework overcomes both limitations by conditioning future generation on recovered trajectories while preserving explicit visibility and correction cues.

\subsection{Learning from Imperfect Historical Conditions}
\label{sec:rw_conditioning}
Mismatches between clean training and imperfect deployment histories have largely been studied under variable observation lengths via distillation~\cite{li2024lakd} or contrastive learning~\cite{qiu2025clls}, while broader time-series works address temporal noise and uncertainty through information bottlenecks~\cite{feng2024timesieve,lai2025faithfultimesieve} or rejection mechanisms~\cite{feng2026reliableforecasting}.
Under ego-view perception, however, historical recovery alters both coordinate estimates and spatial reliability within a fixed window, creating a conditioning shift.
Inspired by curriculum learning~\cite{bengio2009curriculum} and scheduled sampling~\cite{bengio2015scheduled}, we introduce an agent-wise curriculum that transitions conditions from clean histories to recovered estimates, adapting the predictor to residual recovery errors.

\begin{figure*}[t]
\centering
\includegraphics[width=\linewidth]{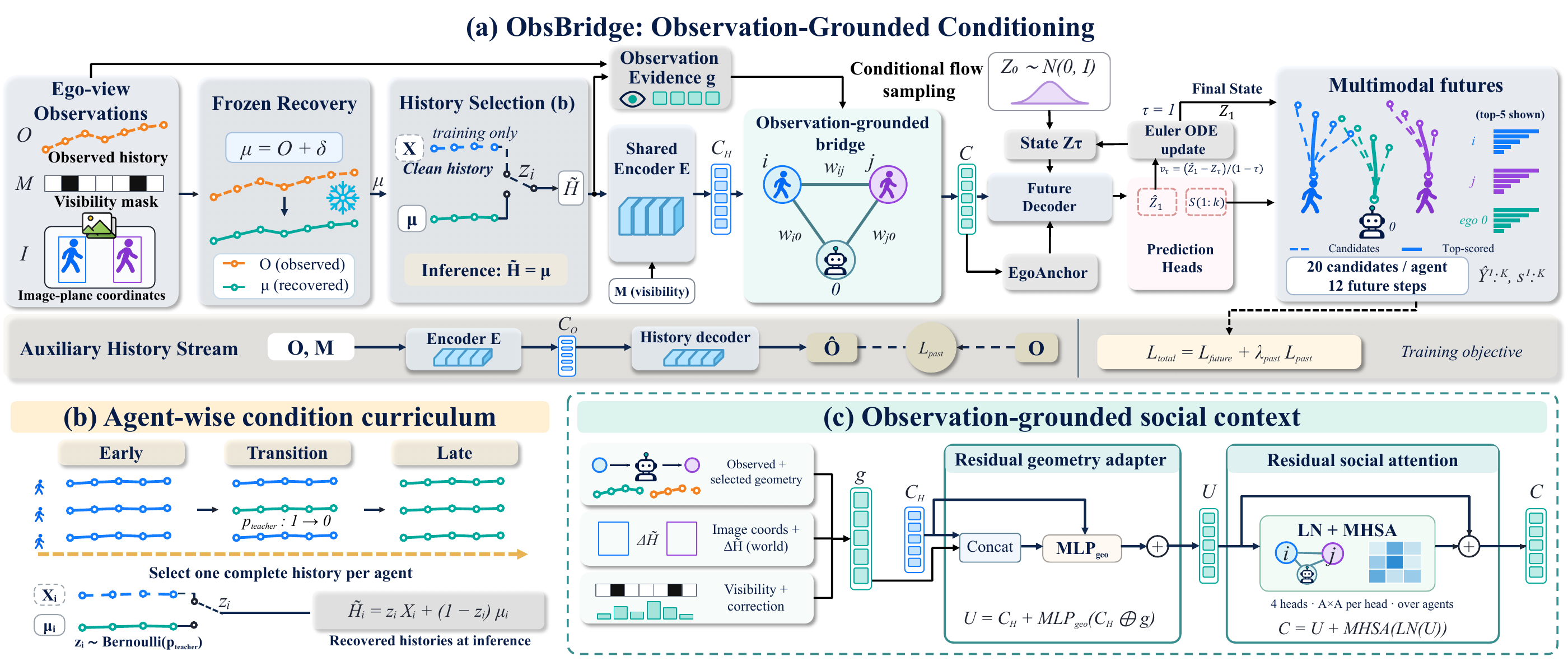}
\caption{\textbf{Overview of the ObsBridge framework.} \textbf{(a)} A frozen estimator recovers historical trajectories $\mu$ from noisy ego-view inputs $(O, M, I)$. The encoded history $C_H$ is augmented with observation evidence $g$ to condition the flow-based future generator alongside an auxiliary history stream. \textbf{(b)} The agent-wise curriculum progressively shifts training conditions $\tilde{H}_i$ from clean ground truth $X_i$ to recovered estimates $\mu_i$. \textbf{(c)} A residual geometry adapter and social attention layer integrate $C_H$ with explicit geometric, visibility, and correction cues $g$ to produce the enhanced context $C$.}
\label{fig:framework}
\label{fig:pipeline}
\end{figure*}

\section{Method}
\label{sec:method}
Existing ego-view trajectory predictors either fail to handle perception noise under corrupted coordinates or discard recovered histories during inference, leaving models vulnerable to distribution shifts.
To resolve this, ObsBridge bridges historical motion recovery and generative forecasting by explicitly propagating denoised states and grounding them in raw observation evidence.
\subsection{Problem Formulation}
\label{sec:method_formulation}
Consider a multi-agent scene with $A$ valid agents, indexed by $i=0$ for the ego agent and $i \in \{1,\ldots, A-1\}$ for surrounding pedestrians.
Over $T_p$ historical steps, the FPV perception yields a noisy observation tuple for each agent $i$:
\begin{equation}
 \mathcal O_i = (O_i, M_i, I_i),
\label{eq:task_input}
\end{equation}
where $O_i \in \mathbb R^{T_p \times 2}$ denotes metric positions, $M_i \in \{0,1\}^{T_p}$ is the binary visibility mask, and $I_i \in \mathbb R^{T_p \times 2}$ represents image-plane coordinates.
Given joint scene observations $\mathcal{O}_{\mathcal{A}} = \{\mathcal O_i\}_{i=0}^{A-1}$, our goal is to learn a noise-resistant predictor that outputs $K$ multimodal future trajectory candidates $\widehat Y_i^{1:K} \in \mathbb R^{K \times T_f \times 2}$ and their mode ranking logits $s_i \in \mathbb R^K$ over the next $T_f$ steps:
\begin{equation}
 \bigl\{(\widehat Y_i^{1:K},\; s_i)\bigr\}_{i=0}^{A-1} = \mathcal F_\Theta\bigl(\mathcal{O}_{\mathcal{A}}\bigr).
\label{eq:task_output}
\end{equation}
In training, clean ground-truth histories $X_i \in \mathbb R^{T_p \times 2}$ and futures $Y_i \in \mathbb R^{T_f \times 2}$ are available for supervision, whereas deployment relies solely on corrupted observations $\mathcal{O}_{\mathcal{A}}$.

\subsection{History Recovery and Condition Curriculum}
\label{sec:method_history}

\textbf{Translation-invariant history recovery.}
To correct noisy pedestrian histories without overfitting to an arbitrary global origin, we construct a translation-invariant input feature $f_i$ for each pedestrian using endpoint-relative displacements, ego-relative displacements, normalized and clipped image coordinates $\eta(\bar I_i)$, and visibility $M_i$:
\begin{equation}
 f_i = (O_i - O_{i,T_p}) \oplus (O_i - O_0) \oplus \eta(\bar I_i) \oplus M_i,
\label{eq:recovery_input}
\end{equation}
where $\oplus$ denotes concatenation of flattened temporal features and $O_{i,T_p}$ is the observed historical endpoint. For image size $W\times H$, $\bar I_{i,t}=(2u_{i,t}/W-1,2v_{i,t}/H-1)$ denotes normalized image coordinates, and $\eta$ clips each coordinate to $[-2,2]$.
An MLP estimator takes $f_i$ to predict historical position corrections $\delta_i \in \mathbb R^{T_p \times 2}$, yielding the recovered trajectory $\mu_i$:
\begin{equation}
 \delta_i = \mathrm{MLP}_{\mathrm{est}}(f_i), \qquad \mu_i = O_i + \delta_i,
\label{eq:recovery_output}
\end{equation}
while the ego agent retains its clean odometry history $\mu_0 = O_0$.
During offline training of the estimator against the clean residual $D_i = X_i - O_i$, the network also predicts an auxiliary log-scale $a_i$ to optimize a heteroscedastic Gaussian loss with a supervised temporal consistency term:
\begin{equation}
\begin{aligned}
  \mathcal L_{\mathrm{est}} = \mathrm{Mean}\!\left(a_i + \frac{(D_i - \delta_i)^2}{2\exp(2a_i)}\right) +  \\ \lambda_{\mathrm{smooth}}\mathrm{SmoothL1}(\Delta\delta_i, \Delta D_i),   
\end{aligned}
\label{eq:estimator_loss}
\end{equation}
where $\Delta$ denotes adjacent-frame differences. After offline training, the estimator is frozen. Only its recovered mean is used by the forecaster, while the auxiliary log-scale $a_i$ is not used as a forecasting input.
To generate training estimates, we perform five-fold cross-fitting over the stored training windows. Validation and test estimates are produced by an estimator trained on all training pedestrians.

\textbf{Agent-wise teacher-to-estimate curriculum.}
To mitigate the conditioning-source shift associated with imperfect estimates $\mu_i$, we apply an agent-wise curriculum over the first $S_c$ training steps.
At step $s$, each agent independently samples its complete historical condition $\widetilde H_i$ between the clean history $X_i$ and the recovered estimate $\mu_i$:
\begin{equation}
 p_{\mathrm{teacher}}(s) = \max\!\left(0,\, 1 - \frac{s}{S_c}\right),
\label{eq:curriculum_prob}
\end{equation}
\begin{equation}
 \widetilde H_i = z_i X_i + (1 - z_i)\mu_i, \qquad z_i \sim \mathrm{Bernoulli}(p_{\mathrm{teacher}}(s)).
\label{eq:curriculum_select}
\end{equation}
This schedule initially favors clean histories and progressively increases exposure to recovered histories.
After $S_c$ steps and during inference, we adopt recovered-history conditioning by setting $\widetilde H_i = \mu_i$.
Following BiFlow~\cite{liu2026egotraj}, a shared contextual encoder then maps both the selected history $\widetilde H$ and the raw observation $O$ into latent features:
\begin{equation}
 C_H = \mathrm{Encoder}(\widetilde H, M), \qquad C_O = \mathrm{Encoder}(O, M),
\label{eq:shared_encoder}
\end{equation}
where $C_H \in \mathbb R^{A \times D}$ serves the future prediction stream and $C_O \in \mathbb R^{A \times D}$ serves the auxiliary history reconstruction stream.

\subsection{Observation-Grounded Social Context}
\label{sec:method_geometry}

The selected historical coordinates $\widetilde H_i$ alone omit how strongly the raw observations support that motion, as similar recovered trajectories can arise from different visibility patterns and correction magnitudes.
To retain observation cues alongside recovered motion, we construct a per-agent geometric evidence descriptor $g_i$ in the world frame by concatenating observed and recovered ego-relative geometry, frame-to-frame displacements $\Delta\widetilde H_i$, normalized image coordinates $\bar I_i$, visibility $M_i$, and correction magnitude:
\begin{equation}
 g_i = (O_i - O_0) \oplus (\widetilde H_i - O_0) \oplus \Delta\widetilde H_i \oplus \bar I_i \oplus M_i \oplus \|\widetilde H_i - O_i\|_2,
\label{eq:geo_evidence}
\end{equation}
augmented with the Euclidean norms of the two ego-relative vectors and a binary ego indicator. Norms are computed framewise, and the final displacement is padded with zero.

We then fuse $g_i$ with the encoded motion $C_H$ through a residual geometry adapter, and aggregate observation-grounded interactions across agents via multi-head self-attention (MHSA):
\begin{equation}
 U = C_H + \mathrm{MLP}_{\mathrm{geo}}(C_H \oplus g),
\label{eq:geo_adapter}
\end{equation}
\begin{equation}
 C = U + \mathrm{MHSA}\bigl(\mathrm{LayerNorm}(U)\bigr),
\label{eq:social_mhsa}
\end{equation}
where $U \in \mathbb R^{A \times D}$ is the observation-grounded node representation and $C \in \mathbb R^{A \times D}$ is the enhanced future context.
The output layers of both $\mathrm{MLP}_{\mathrm{geo}}$ and $\mathrm{MHSA}$ are zero-initialized, so the bridging module starts as an identity refinement ($C = C_H$) and progressively injects observation-grounded social context during training.

\begin{table*}[t]
\centering
\caption{Quantitative results  using minADE@20 / minFDE@20 (in meters; lower is better).
The better result in the final block is highlighted in \textbf{bold}, while the
\protect\colorbox{obrankone}{best},
\protect\colorbox{obranktwo}{second-best}, and
\protect\colorbox{obrankthree}{third-best} distinct values in each metric column are highlighted by background colors. For BiFlow, we evaluated with the author's latest EgoTraj-TBD dataset version and its corrected results.}
\label{tab:main_comparison}
\footnotesize
\setlength{\tabcolsep}{2pt}
\renewcommand{\arraystretch}{0.9} 
\setlength{\extrarowheight}{0pt}
\setlength{\aboverulesep}{0pt}
\setlength{\belowrulesep}{0pt}
\setbox\strutbox=\hbox{\vrule height8.5pt depth3.5pt width0pt}
\begin{tabularx}{\textwidth}{@{}c
*{2}{>{\hsize=1.18\hsize\linewidth=\hsize\centering\arraybackslash}X}
*{5}{>{\hsize=.97\hsize\linewidth=\hsize\centering\arraybackslash}X>{\hsize=.97\hsize\linewidth=\hsize\centering\arraybackslash}X}
>{\hsize=.97\hsize\linewidth=\hsize\centering\arraybackslash}X>{\hsize=.97\hsize\linewidth=\hsize\centering\arraybackslash}X}
\toprule
\multirow{2}{*}{\textbf{Method}}
 & \multicolumn{2}{c}{\multirow{2}{*}{\textbf{EgoTraj-TBD}~\cite{liu2026egotraj}}}
 & \multicolumn{12}{c@{}}{\textbf{T2FPV-ETH}~\cite{stoler2023t2fpv}} \\
\cline{4-15}
 & \multicolumn{2}{c}{}
 & \multicolumn{2}{c}{ETH}
 & \multicolumn{2}{c}{HOTEL}
 & \multicolumn{2}{c}{ZARA1}
 & \multicolumn{2}{c}{ZARA2}
 & \multicolumn{2}{c}{UNIV}
 & \multicolumn{2}{c@{}}{\textbf{AVG}} \\
\midrule
VRNN~\cite{chung2015recurrent} & 0.76 & 1.26 & 1.35 & 2.00 & 1.30 & 1.73 & 1.14 & 1.68 & 1.54 & 2.10 & 2.24 & 2.89 & 1.51 & 2.08 \\
VRNN + CoFE & 0.68 & 1.19 & 1.52 & 2.35 & 1.06 & 1.53 & 1.65 & 2.10 & 1.06 & 1.63 & 1.27 & 1.62 & 1.31 & 1.85 \\
AC-VRNN~\cite{bertugli2021ac} & 0.63 & 1.00 & 1.39 & 2.04 & 1.31 & 1.75 & 1.04 & 1.40 & 1.47 & 1.93 & 2.26 & 3.00 & 1.49 & 2.02 \\
AC-VRNN + CoFE & 0.53 & 0.85 & 1.47 & 2.18 & 1.16 & 1.72 & 1.03 & 1.48 & 1.31 & 1.69 & 1.54 & 1.88 & 1.30 & 1.79 \\
SGNet~\cite{wang2022stepwise} & 0.40 & 0.59 & 1.43 & 1.97 & 0.72 & 1.00 & 0.58 & 0.79 & 0.78 & 0.91 & 1.48 & 1.73 & 1.00 & 1.28 \\
SGNet + CoFE & 0.37 & 0.53 & 0.98 & 1.32 & 0.59 & 0.76 & 0.55 & 0.76 & 0.73 & 0.86 & 1.23 & 1.48 & 0.82 & 1.04 \\
TUTR*~\cite{shi2023trajectory} & 0.37 & 0.59 & 1.25 & 1.54 & 1.04 & 1.47 & 0.77 & 1.00 & 0.75 & 0.91 & 1.10 & 1.30 & 0.98 & 1.24 \\
TUTR* + CoFE & 0.38 & 0.60 & 1.02 & 1.41 & 0.85 & 1.26 & 0.62 & 0.93 & 0.72 & 0.82 & 1.02 & 1.24 & 0.85 & 1.13 \\
MoFlow~\cite{fu2025moflow} & \thirdcell{0.32} & \secondcell{0.45} & \thirdcell{0.85} & 1.22 & 0.63 & 0.89 & \thirdcell{0.46} & \thirdcell{0.63} & \bestcell{0.50} & \bestcell{0.60} & \secondcell{0.92} & \thirdcell{1.10} & \thirdcell{0.67} & \thirdcell{0.89} \\
MoFlow + CoFE & \secondcell{0.31} & \secondcell{0.45} & 0.88 & 1.26 & 0.62 & 0.87 & 0.52 & 0.67 & \thirdcell{0.53} & \thirdcell{0.63} & \secondcell{0.92} & \secondcell{1.08} & 0.69 & 0.90 \\
\midrule
Resonance~\cite{wong2025resonance} & 0.38 & 0.56 & 0.88 & \thirdcell{1.19} & \thirdcell{0.57} & 0.78 & 0.50 & 0.70 & 0.64 & 0.83 & 1.41 & 1.96 & 0.80 & 1.09 \\
INTraJ~\cite{chen2026intraj} & 0.37 & 0.55 & 0.86 & 1.21 & 0.58 & 0.81 & 0.49 & 0.70 & 0.61 & 0.77 & 1.39 & 1.91 & 0.79 & 1.08 \\
Social-Mamba~\cite{luan2026social} & \thirdcell{0.32} & 0.48 & 0.96 & 1.34 & 0.60 & 0.82 & 0.49 & 0.66 & \thirdcell{0.53} & 0.64 & 0.99 & 1.18 & 0.71 & 0.93 \\
DD-MDN~\cite{hetzel2026ddmdn} & 0.35 & 0.53 & 1.17 & 1.68 & \bestcell{0.39} & \bestcell{0.47} & 0.49 & 0.71 & 0.56 & 0.70 & \thirdcell{0.97} & 1.16 & 0.72 & 0.94 \\
\midrule
BiFlow~\cite{liu2026egotraj} & \secondcell{0.31} & \thirdcell{0.46} & \secondcell{0.74} & \secondcell{1.07} & 0.59 & \thirdcell{0.73} & \secondcell{0.43} & \secondcell{0.58} & \secondcell{0.51} & \secondcell{0.61} & \secondcell{0.92} & \thirdcell{1.10} & \secondcell{0.64} & \secondcell{0.82} \\
\textbf{ObsBridge (ours)} & \bestcell{\textbf{0.29}} & \bestcell{\textbf{0.42}} & \bestcell{\textbf{0.70}} & \bestcell{\textbf{1.01}} & \secondcell{\textbf{0.50}} & \secondcell{\textbf{0.60}} & \bestcell{\textbf{0.41}} & \bestcell{\textbf{0.54}} & \bestcell{\textbf{0.50}} & \bestcell{\textbf{0.60}} & \bestcell{\textbf{0.89}} & \bestcell{\textbf{1.07}} & \bestcell{\textbf{0.60}} & \bestcell{\textbf{0.76}} \\
\bottomrule
\end{tabularx}
\par\vspace{2pt}
\parbox{\textwidth}{%
    \footnotesize\raggedright
    $^*$ The TUTR architecture is adapted for multimodal prediction following
    EgoTraj-Bench~\cite{liu2026egotraj}.\par
}
\end{table*}
\subsection{Dual-Stream Flow Matching and Optimization}
\label{sec:method_training}

We feed the enhanced context $C$ into the future prediction decoder and EgoAnchor of BiFlow~\cite{liu2026egotraj}, while routing $C_O$ to the history reconstruction decoder.
Following conditional flow matching~\cite{lipman2023flowmatching}, let $Z_1=\Psi_O(Y)$ denote the clean future after BiFlow's observation-derived rotation, observed-endpoint anchoring, and min--max normalization. For Gaussian noise $Z_0\sim\mathcal N(0,I)$, the interpolated state at flow time $\tau\in[0,1)$ is:
\begin{equation}
 Z_\tau = (1 - \tau)Z_0 + \tau Z_1.
\label{eq:flow_interp}
\end{equation}
Conditioned on $C$ and $Z_\tau$, the future decoder predicts $K$ endpoint states $\widehat Z_1^{1:K}$ and per-agent mode logits $s_i^{1:K}$, yielding metric trajectories $\widehat Y^k=\Psi_O^{-1}(\widehat Z_1^k)$. The candidate slots share each agent's base Gaussian initialization and are distinguished by learned hypothesis embeddings.
For each agent $i$, define the time-summed position error $d_i^k=\sum_{t=1}^{T_f}\|\widehat Y_{i,t}^k-Y_{i,t}\|_2$.
The future prediction objective supervises the best-matching candidate $k_i^*$ and mode selection:
\begin{equation}
\begin{aligned}
 \mathcal L_{\mathrm{future}}
 &= \mathbb E\!\left[\frac{1}{A}\sum_{i=0}^{A-1}
 \left(d_i^{k_i^*}+\mathrm{CE}(s_i^{1:K},k_i^*)\right)\right],\\
 k_i^* &= \operatorname*{arg\,min}_{k} d_i^k,
\end{aligned}
\label{eq:future_loss}
\end{equation}
where $\mathrm{CE}(\cdot)$ is the cross-entropy loss and the expectation averages over training scenes, flow times, and Gaussian noise.
Following BiFlow~\cite{liu2026egotraj}, we retain the auxiliary historical flow-matching objective $\mathcal{L}_{\mathrm{past}}$.
With the historical estimator frozen, the forecasting network is trained via:
\begin{equation}
 \mathcal L_{\mathrm{total}} = \mathcal L_{\mathrm{future}} + \lambda_{\mathrm{past}}\mathcal L_{\mathrm{past}}.
\label{eq:total_loss}
\end{equation}
During inference, the inherited BiFlow sampler evaluates both streams, while only future trajectories and their mode logits are used as forecasting outputs. Starting from the shared Gaussian initialization, the future states are iteratively updated through $v_\theta^k(Z_\tau,\tau;C)=(\widehat Z_1^k-Z_\tau^k)/(1-\tau)$ and mapped back to metric coordinates using $\Psi_O^{-1}$.

\label{sec:experiments}
\section{Experiments}
\label{sec:experiments}

\subsection{Experimental Setup}
\label{sec:exp_setup}
We evaluate on the real-world EgoTraj-TBD~\cite{liu2026egotraj} and synthetic T2FPV-ETH~\cite{stoler2023t2fpv} (five splits) benchmarks. Following standard protocols~\cite{liu2026egotraj}, models take $T_p=8$ historical frames (3.2\,s) to predict $T_f=12$ future frames (4.8\,s) with $K=20$ candidates, evaluated via ADE/FDE at @1 and minADE/minFDE at @5 and @20 (in meters). We train the recovery estimator for 80 epochs ($\lambda_{\text{smooth}}=0.1$) and the forecaster for 150 epochs using AdamW (batch size 32, learning rate $10^{-4}$), with a 50-epoch curriculum and $\lambda_{\text{past}}=0.5$. 
All other training protocols and test-time ODE sampling settings strictly match BiFlow~\cite{liu2026egotraj}. Further implementation details are provided in the supplementary material.

\begin{table*}[t]
\centering
\definecolor{obdeltablue}{HTML}{2E75B6}
\caption{Quantitative results across EgoTraj-TBD~\cite{liu2026egotraj} and T2FPV-ETH~\cite{stoler2023t2fpv} using ADE/FDE at @1 and minADE/minFDE at @5 and @20 (in meters; lower is better).
The best result is highlighted in \textbf{bold}
(\protect\textcolor{obdeltablue}{$\blacktriangledown$}: decrease;
\protect\textcolor{obdeltablue}{$\blacktriangle$}: increase).}
\label{tab:ranked_comparison}
\footnotesize
\newcommand{\obdelta}[2]{\mbox{#1\raisebox{-0.7pt}{\normalfont\fontsize{6.5pt}{7pt}\selectfont\color{obdeltablue}#2}}}
\setlength{\tabcolsep}{2pt}
\renewcommand{\arraystretch}{1.0} 
\setlength{\extrarowheight}{0pt}
\setlength{\aboverulesep}{0pt}
\setlength{\belowrulesep}{0pt}
\setbox\strutbox=\hbox{\vrule height8.5pt depth3.5pt width0pt}
\begin{tabularx}{\textwidth}{c
*{2}{>{\hsize=.85\hsize\linewidth=\hsize\centering\arraybackslash}X>{\hsize=.85\hsize\linewidth=\hsize\centering\arraybackslash}X
>{\hsize=1.15\hsize\linewidth=\hsize\centering\arraybackslash}X>{\hsize=1.15\hsize\linewidth=\hsize\centering\arraybackslash}X}
>{\hsize=.85\hsize\linewidth=\hsize\centering\arraybackslash}X>{\hsize=.85\hsize\linewidth=\hsize\centering\arraybackslash}X
>{\hsize=1.15\hsize\linewidth=\hsize\centering\arraybackslash}X>{\hsize=1.15\hsize\linewidth=\hsize\centering\arraybackslash}X}
\toprule
\multirow{2}{*}{Dataset}
 & \multicolumn{4}{c}{\textbf{@1}}
 & \multicolumn{4}{c}{\textbf{@5}}
 & \multicolumn{4}{c}{\textbf{@20}} \\
\cline{2-13}
 & \multicolumn{2}{c}{BiFlow~\cite{liu2026egotraj}}
 & \multicolumn{2}{c}{\textbf{ObsBridge}}
 & \multicolumn{2}{c}{BiFlow~\cite{liu2026egotraj}}
 & \multicolumn{2}{c}{\textbf{ObsBridge}}
 & \multicolumn{2}{c}{BiFlow~\cite{liu2026egotraj}}
 & \multicolumn{2}{c}{\textbf{ObsBridge}} \\
\midrule
EgoTraj-TBD & 0.98 & 1.69 & \obdelta{\textbf{0.87}}{\ensuremath{\blacktriangledown}11.2\%} & \obdelta{\textbf{1.47}}{\ensuremath{\blacktriangledown}13.0\%} & 0.60 & 1.02 & \obdelta{\textbf{0.55}}{\ensuremath{\blacktriangledown}8.3\%} & \obdelta{\textbf{0.92}}{\ensuremath{\blacktriangledown}9.8\%} & 0.31 & 0.46 & \obdelta{\textbf{0.29}}{\ensuremath{\blacktriangledown}6.5\%} & \obdelta{\textbf{0.42}}{\ensuremath{\blacktriangledown}8.7\%} \\
\midrule
\rowcolor{black!6}
ETH & 1.90 & 3.05 & \obdelta{\textbf{1.47}}{\ensuremath{\blacktriangledown}22.6\%} & \obdelta{\textbf{2.50}}{\ensuremath{\blacktriangledown}18.0\%} & 1.43 & 2.34 & \obdelta{\textbf{1.16}}{\ensuremath{\blacktriangledown}18.9\%} & \obdelta{\textbf{1.93}}{\ensuremath{\blacktriangledown}17.5\%} & 0.74 & 1.07 & \obdelta{\textbf{0.70}}{\ensuremath{\blacktriangledown}5.4\%} & \obdelta{\textbf{1.01}}{\ensuremath{\blacktriangledown}5.6\%} \\
HOTEL & 1.77 & 2.53 & \obdelta{\textbf{1.29}}{\ensuremath{\blacktriangledown}27.1\%} & \obdelta{\textbf{1.88}}{\ensuremath{\blacktriangledown}25.7\%} & 1.17 & 1.65 & \obdelta{\textbf{0.88}}{\ensuremath{\blacktriangledown}24.8\%} & \obdelta{\textbf{1.20}}{\ensuremath{\blacktriangledown}27.3\%} & 0.59 & 0.73 & \obdelta{\textbf{0.50}}{\ensuremath{\blacktriangledown}15.3\%} & \obdelta{\textbf{0.60}}{\ensuremath{\blacktriangledown}17.8\%} \\
\rowcolor{black!6}
ZARA1 & 1.25 & 2.02 & \obdelta{\textbf{1.20}}{\ensuremath{\blacktriangledown}4.0\%} & \obdelta{\textbf{1.96}}{\ensuremath{\blacktriangledown}3.0\%} & 0.81 & \textbf{1.26} & \obdelta{\textbf{0.81 }}{0.0\%} & \obdelta{1.27}{\ensuremath{\blacktriangle}0.8\%} & 0.43 & 0.58 & \obdelta{\textbf{0.41}}{\ensuremath{\blacktriangledown}4.7\%} & \obdelta{\textbf{0.54}}{\ensuremath{\blacktriangledown}6.9\%} \\
ZARA2 & 1.43 & 2.10 & \obdelta{\textbf{1.39}}{\ensuremath{\blacktriangledown}2.8\%} & \obdelta{\textbf{2.06}}{\ensuremath{\blacktriangledown}1.9\%} & 0.99 & 1.38 & \obdelta{\textbf{0.97}}{\ensuremath{\blacktriangledown}2.0\%} & \obdelta{\textbf{1.36}}{\ensuremath{\blacktriangledown}1.4\%} & 0.51 & 0.61 & \obdelta{\textbf{0.50}}{\ensuremath{\blacktriangledown}2.0\%} & \obdelta{\textbf{0.60}}{\ensuremath{\blacktriangledown}1.6\%} \\
\rowcolor{black!6}
UNIV & 2.38 & 3.44 & \obdelta{\textbf{2.26}}{\ensuremath{\blacktriangledown}5.0\%} & \obdelta{\textbf{3.20}}{\ensuremath{\blacktriangledown}7.0\%} & 1.75 & 2.43 & \obdelta{\textbf{1.70}}{\ensuremath{\blacktriangledown}2.9\%} & \obdelta{\textbf{2.31}}{\ensuremath{\blacktriangledown}4.9\%} & 0.92 & 1.10 & \obdelta{\textbf{0.89}}{\ensuremath{\blacktriangledown}3.3\%} & \obdelta{\textbf{1.07}}{\ensuremath{\blacktriangledown}2.7\%} \\
\midrule
AVG & 1.75 & 2.63 & \obdelta{\textbf{1.52}}{\ensuremath{\blacktriangledown}13.1\%} & \obdelta{\textbf{2.32}}{\ensuremath{\blacktriangledown}11.8\%} & 1.23 & 1.81 & \obdelta{\textbf{1.10}}{\ensuremath{\blacktriangledown}10.6\%} & \obdelta{\textbf{1.61}}{\ensuremath{\blacktriangledown}11.0\%} & 0.64 & 0.82 & \obdelta{\textbf{0.60}}{\ensuremath{\blacktriangledown}6.3\%} & \obdelta{\textbf{0.76}}{\ensuremath{\blacktriangledown}7.3\%} \\
\bottomrule
\end{tabularx}
\end{table*}
\begin{figure*}[t]
    \centering
    \includegraphics[width=\textwidth]{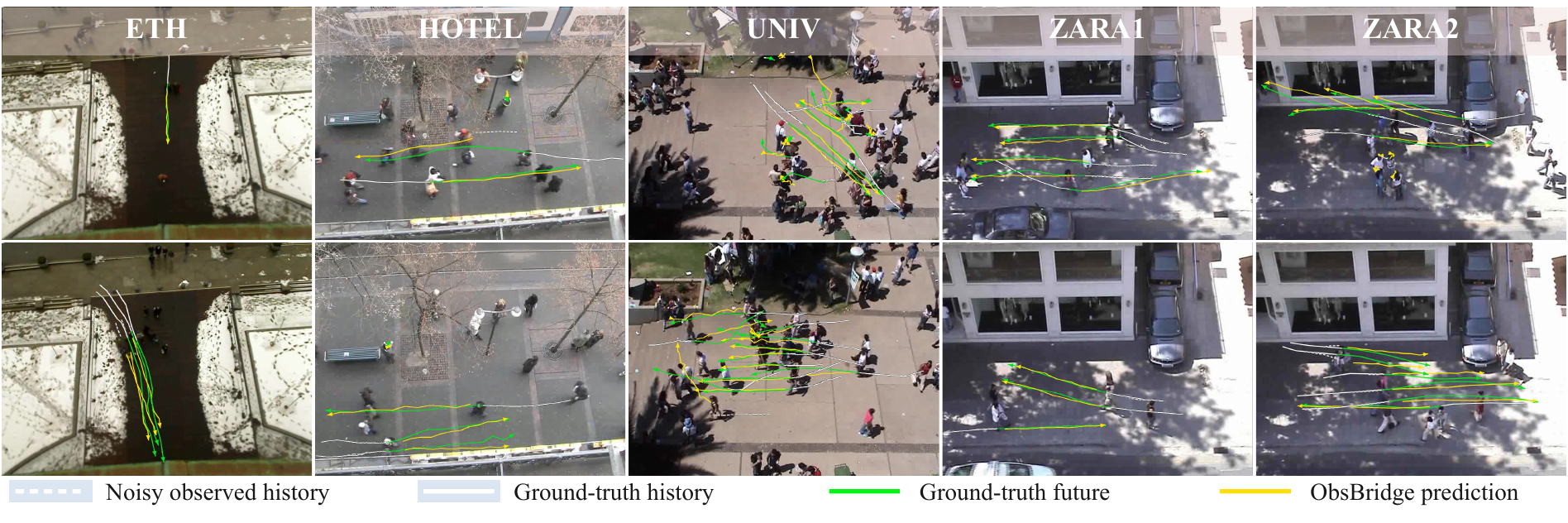}
    \caption{Qualitative results on T2FPV-ETH.
    The predicted future trajectories of ObsBridge closely align with the ground-truth futures across the five splits, capturing diverse pedestrian motion patterns under noisy ego-view observations.}
    \label{fig:qualitative_t2fpv}
\end{figure*}

\begin{figure*}[t]
    \centering
    \includegraphics[width=\textwidth]{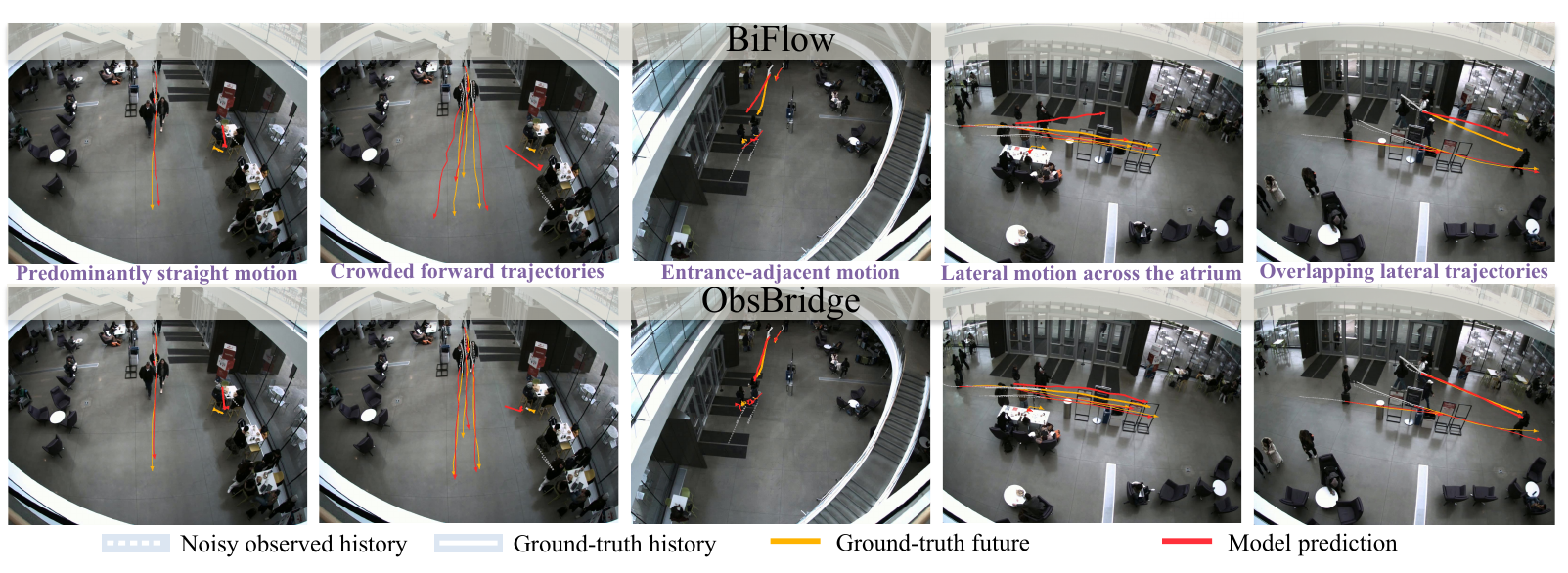}
    \caption{Qualitative comparison between BiFlow (top) and ObsBridge (bottom) on EgoTraj-TBD.}
    \label{fig:qualitative_tbd}
\end{figure*}


\begin{table*}[t]
\centering
\definecolor{obdeltablue}{HTML}{2E75B6}
\caption{\textbf{Results across different numbers of valid agents on EgoTraj-TBD}~\cite{liu2026egotraj}. \textbf{Bold} indicates the lower error, and \protect\colorbox{obrankone}{blue} annotations denote relative changes over BiFlow (\protect\textcolor{obdeltablue}{$\blacktriangledown$}: decrease; \protect\textcolor{obdeltablue}{$\blacktriangle$}: increase). $N$ denotes the number of valid objects per window, including ego.  Scenes denotes the number of evaluated scenes in each group. Each group is evaluated over all valid objects.}
\label{tab:object_count}
\footnotesize
\newcommand{\obcountdelta}[2]{\mbox{#1\raisebox{-0.7pt}{\normalfont\fontsize{6.5pt}{7pt}\selectfont\color{obdeltablue}#2}}}
\setlength{\tabcolsep}{2pt}
\renewcommand{\arraystretch}{1.0} 
\setlength{\extrarowheight}{0pt}
\setlength{\aboverulesep}{0pt}
\setlength{\belowrulesep}{0pt}
\setbox\strutbox=\hbox{\vrule height8.5pt depth3.5pt width0pt}
\begin{tabularx}{\textwidth}{cc
*{3}{
>{\hsize=.85\hsize\linewidth=\hsize\centering\arraybackslash}X
>{\hsize=.85\hsize\linewidth=\hsize\centering\arraybackslash}X
>{\hsize=1.15\hsize\linewidth=\hsize\centering\arraybackslash}X
>{\hsize=1.15\hsize\linewidth=\hsize\centering\arraybackslash}X
}}
\toprule
\multirow{2}{*}{\textbf{$N$}}
& \multirow{2}{*}{\textbf{Scenes}}
& \multicolumn{4}{c}{\textbf{@1}}
& \multicolumn{4}{c}{\textbf{@5}}
& \multicolumn{4}{c}{\textbf{@20}} \\
\cline{3-14}
& 
& \multicolumn{2}{c}{BiFlow}
& \multicolumn{2}{c}{\textbf{ObsBridge}}
& \multicolumn{2}{c}{BiFlow}
& \multicolumn{2}{c}{\textbf{ObsBridge}}
& \multicolumn{2}{c}{BiFlow}
& \multicolumn{2}{c}{\textbf{ObsBridge}} \\
\midrule
1 & 5,378
& 0.60 & 1.28
& \obcountdelta{\textbf{0.56}}{\ensuremath{\blacktriangledown}6.7\%}
& \obcountdelta{\textbf{1.16}}{\ensuremath{\blacktriangledown}9.4\%}
& 0.39 & 0.80
& \obcountdelta{\textbf{0.36}}{\ensuremath{\blacktriangledown}7.7\%}
& \obcountdelta{\textbf{0.73}}{\ensuremath{\blacktriangledown}8.8\%}
& 0.21 & 0.35
& \obcountdelta{\textbf{0.19}}{\ensuremath{\blacktriangledown}9.5\%}
& \obcountdelta{\textbf{0.31}}{\ensuremath{\blacktriangledown}11.4\%} \\
\rowcolor{black!6}
2 & 744
& 1.26 & 2.04
& \obcountdelta{\textbf{1.18}}{\ensuremath{\blacktriangledown}6.3\%}
& \obcountdelta{\textbf{1.86}}{\ensuremath{\blacktriangledown}8.8\%}
& 0.77 & 1.22
& \obcountdelta{\textbf{0.72}}{\ensuremath{\blacktriangledown}6.5\%}
& \obcountdelta{\textbf{1.12}}{\ensuremath{\blacktriangledown}8.2\%}
& 0.40 & 0.57
& \obcountdelta{\textbf{0.38}}{\ensuremath{\blacktriangledown}5.0\%}
& \obcountdelta{\textbf{0.54}}{\ensuremath{\blacktriangledown}5.3\%} \\
3 & 434
& 1.39 & 2.10
& \obcountdelta{\textbf{1.17}}{\ensuremath{\blacktriangledown}15.8\%}
& \obcountdelta{\textbf{1.75}}{\ensuremath{\blacktriangledown}16.7\%}
& 0.79 & 1.17
& \obcountdelta{\textbf{0.72}}{\ensuremath{\blacktriangledown}8.9\%}
& \obcountdelta{\textbf{1.07}}{\ensuremath{\blacktriangledown}8.5\%}
& \textbf{0.38} & 0.50
& \obcountdelta{0.39}{\ensuremath{\blacktriangle}2.6\%}
& \obcountdelta{0.50}{0.0\%} \\
\rowcolor{black!6}
4 & 205
& 1.55 & 2.30
& \obcountdelta{\textbf{1.27}}{\ensuremath{\blacktriangledown}18.1\%}
& \obcountdelta{\textbf{1.84}}{\ensuremath{\blacktriangledown}20.0\%}
& 0.87 & 1.28
& \obcountdelta{\textbf{0.78}}{\ensuremath{\blacktriangledown}10.3\%}
& \obcountdelta{\textbf{1.13}}{\ensuremath{\blacktriangledown}11.7\%}
& 0.43 & 0.57
& \obcountdelta{\textbf{0.41}}{\ensuremath{\blacktriangledown}4.7\%}
& \obcountdelta{\textbf{0.53}}{\ensuremath{\blacktriangledown}7.0\%} \\
$\geq 5$ & 147
& 1.71 & 2.46
& \obcountdelta{\textbf{1.40}}{\ensuremath{\blacktriangledown}18.1\%}
& \obcountdelta{\textbf{1.99}}{\ensuremath{\blacktriangledown}19.1\%}
& 1.08 & 1.56
& \obcountdelta{\textbf{0.91}}{\ensuremath{\blacktriangledown}15.7\%}
& \obcountdelta{\textbf{1.31}}{\ensuremath{\blacktriangledown}16.0\%}
& 0.55 & 0.74
& \obcountdelta{\textbf{0.48}}{\ensuremath{\blacktriangledown}12.7\%}
& \obcountdelta{\textbf{0.62}}{\ensuremath{\blacktriangledown}16.2\%} \\
\midrule
All & 6,908
& 0.98 & 1.69
& \obcountdelta{\textbf{0.87}}{\ensuremath{\blacktriangledown}11.2\%}
& \obcountdelta{\textbf{1.47}}{\ensuremath{\blacktriangledown}13.0\%}
& 0.60 & 1.02
& \obcountdelta{\textbf{0.55}}{\ensuremath{\blacktriangledown}8.3\%}
& \obcountdelta{\textbf{0.92}}{\ensuremath{\blacktriangledown}9.8\%}
& 0.31 & 0.46
& \obcountdelta{\textbf{0.29}}{\ensuremath{\blacktriangledown}6.5\%}
& \obcountdelta{\textbf{0.42}}{\ensuremath{\blacktriangledown}8.7\%} \\
\bottomrule
\end{tabularx}
\end{table*}

\subsection{Comparison with the SOTA methods}
\label{sec:exp_main}

Table~\ref{tab:main_comparison} compares ObsBridge with prior trajectory forecasting methods on EgoTraj-TBD and T2FPV-ETH.
Baseline results in the first block are taken from EgoTraj-Bench~\cite{liu2026egotraj}.
For the direct comparison with BiFlow, we retrain the model using its original training procedure and evaluate it alongside ObsBridge with the same test-time sampler.

ObsBridge achieves the lowest minADE@20 and minFDE@20 on EgoTraj-TBD and the lowest average errors on T2FPV-ETH among methods in Table~\ref{tab:main_comparison}.
Compared with our BiFlow reproduction, it improves both metrics across all six test sets.
On EgoTraj-TBD, minADE@20/minFDE@20 decrease from $0.31/0.46$ to $0.29/0.42$\,m, corresponding to reductions of $6.5\%/8.7\%$.
On T2FPV-ETH, average errors decrease from $0.64/0.82$ to $0.60/0.76$\,m, yielding reductions of $6.3\%/7.3\%$.
The largest relative improvements over BiFlow occur on HOTEL, at $15.3\%/17.8\%$.
Figure~\ref{fig:qualitative_t2fpv} visualizes ObsBridge predictions across the five T2FPV-ETH splits.
These gains across real and simulated ego-view observations support the effectiveness of observation-grounded conditioning.


\textbf{Accuracy of highly ranked forecasts.}
Table~\ref{tab:ranked_comparison} evaluates the same 20 hypotheses from each model at three ranking cutoffs.
ObsBridge improves ADE@1 and FDE@1 across all six test sets.
On EgoTraj-TBD, the top-ranked errors decrease from $0.98/1.69$ to $0.87/1.47$\,m, corresponding to reductions of $11.2\%/13.0\%$.
Across T2FPV-ETH, the average top-ranked errors decrease from $1.75/2.63$ to $1.52/2.32$\,m, corresponding to reductions of $13.1\%/11.8\%$, while the average minADE@5/minFDE@5 decrease from $1.23/1.81$ to $1.10/1.61$\,m.
Figure~\ref{fig:qualitative_tbd} further illustrates the improved trajectory predictions of ObsBridge over BiFlow in representative EgoTraj-TBD scenes.
The gains therefore extend from best-of-20 accuracy to the single forecast selected by the model's own score.

\textbf{Performance across different numbers of objects.}
Table~\ref{tab:object_count} further evaluates prediction accuracy under varying numbers of objects on EgoTraj-TBD. ObsBridge consistently improves ADE@1 and FDE@1 across all groups, with more pronounced gains in multi-object scenes.
For scenes containing three and four objects, ADE@1/FDE@1 decrease by $15.8\%/16.7\%$ and $18.1\%/20.0\%$, respectively.
For scenes with at least five objects, the reductions reach $18.1\%/19.1\%$ at @1, $15.7\%/16.0\%$ at @5, and $12.7\%/16.2\%$ at @20.
These results highlight the advantage of ObsBridge in producing accurate high-ranked predictions in multi-object ego-view scenes.

\subsection{Ablation Studies}
\label{sec:exp_ablation}

\textbf{History conditioning.}
Table~\ref{tab:tbd_core_ablation} examines recovered-history conditioning and curriculum adaptation on EgoTraj-TBD.
Compared with our BiFlow reproduction, conditioning on recovered histories throughout training (R) reduces ADE@1/FDE@1 from $0.977/1.691$ to $0.944/1.616$\,m and minADE@5/minFDE@5 from $0.600/1.020$ to $0.588/0.996$\,m.
These gains support incorporating recovered motion into the forecasting condition.
To evaluate curriculum adaptation, we keep the frozen recovery estimator, training and validation protocols, and auxiliary history objective fixed across ObsBridge variants.
Adding the curriculum (R+C) reduces minFDE@20 from $0.457$\,m to $0.449$\,m, while slightly increasing @1 and @5 errors.

\begin{table}[t]
\centering
\caption{Component ablation on EgoTraj-TBD~\cite{liu2026egotraj} dataset. Module R, C, A, and G denote recovered-history conditioning, teacher-to-estimate curriculum, social attention, and residual geometry adapter, respectively.}
\label{tab:tbd_core_ablation}
\footnotesize
\setlength{\tabcolsep}{0.75pt}
\renewcommand{\arraystretch}{1.0} 
\makeatletter
\newcommand{\obgrayrow}{%
  \smash{\rlap{\begingroup%
    \setbox0=\hbox{BiFlow~\cite{liu2026egotraj}}%
    \setbox2=\hbox{ObsBridge}%
    \dimen0=\wd0\relax%
    \ifdim\wd2>\dimen0 \dimen0=\wd2\relax\fi%
    \dimen2=\dimen0\relax%
    \divide\dimen2 by2\relax%
    \advance\dimen2 by\tabcolsep%
    \advance\dimen0 by\tabcolsep%
    \edef\obrowoffset{\the\dimen2}%
    \edef\obrowwidth{\the\dimexpr\columnwidth-\dimen0\relax}%
    \kern\obrowoffset\relax%
    \color{black!6}%
    \rule[-\dp\@arstrutbox]{\obrowwidth}%
         {\dimexpr\ht\@arstrutbox+\dp\@arstrutbox\relax}%
  \endgroup}}%
}
\makeatother
\begin{tabular*}{\columnwidth}{@{\extracolsep{\fill}}ccccc@{\hspace{3pt}}ccc@{}}
\toprule
\multirow{2}{*}{\textbf{Model}} & \multicolumn{4}{c}{\textbf{Modules}} & \multicolumn{3}{c@{}}{\textbf{Displacement error (m)}} \\
\cmidrule(lr){2-5}\cmidrule(lr){6-8}
 & R & C & A & G & \multicolumn{1}{c}{@1} & \multicolumn{1}{c}{@5} & \multicolumn{1}{c@{}}{@20} \\
\midrule
BiFlow & {\color{ourred}\xmark} & {\color{ourred}\xmark} & {\color{ourred}\xmark} & {\color{ourred}\xmark} & 0.977/1.691 & 0.600/1.020 & 0.307/0.458 \\
\midrule
\obgrayrow & {\color{ourgreen}\cmark} & {\color{ourred}\xmark} & {\color{ourred}\xmark} & {\color{ourred}\xmark} & 0.944/1.616 & 0.588/0.996 & 0.310/0.457 \\
 & {\color{ourgreen}\cmark} & {\color{ourgreen}\cmark} & {\color{ourred}\xmark} & {\color{ourred}\xmark} & 0.967/1.657 & 0.594/0.997 & 0.307/0.449 \\
\obgrayrow & {\color{ourgreen}\cmark} & {\color{ourgreen}\cmark} & {\color{ourgreen}\cmark} & {\color{ourred}\xmark} & 0.923/1.593 & 0.597/1.004 & 0.307/0.448 \\
\multirow{-4}{*}{ObsBridge} & {\color{ourgreen}\cmark} & {\color{ourgreen}\cmark} & {\color{ourgreen}\cmark} & {\color{ourgreen}\cmark} & \textbf{0.866}/\textbf{1.470} & \textbf{0.547}/\textbf{0.916} & \textbf{0.289}/\textbf{0.416} \\
\bottomrule
\end{tabular*}
\end{table}

\begin{figure*}[t]
    \centering
    \includegraphics[width=\textwidth]{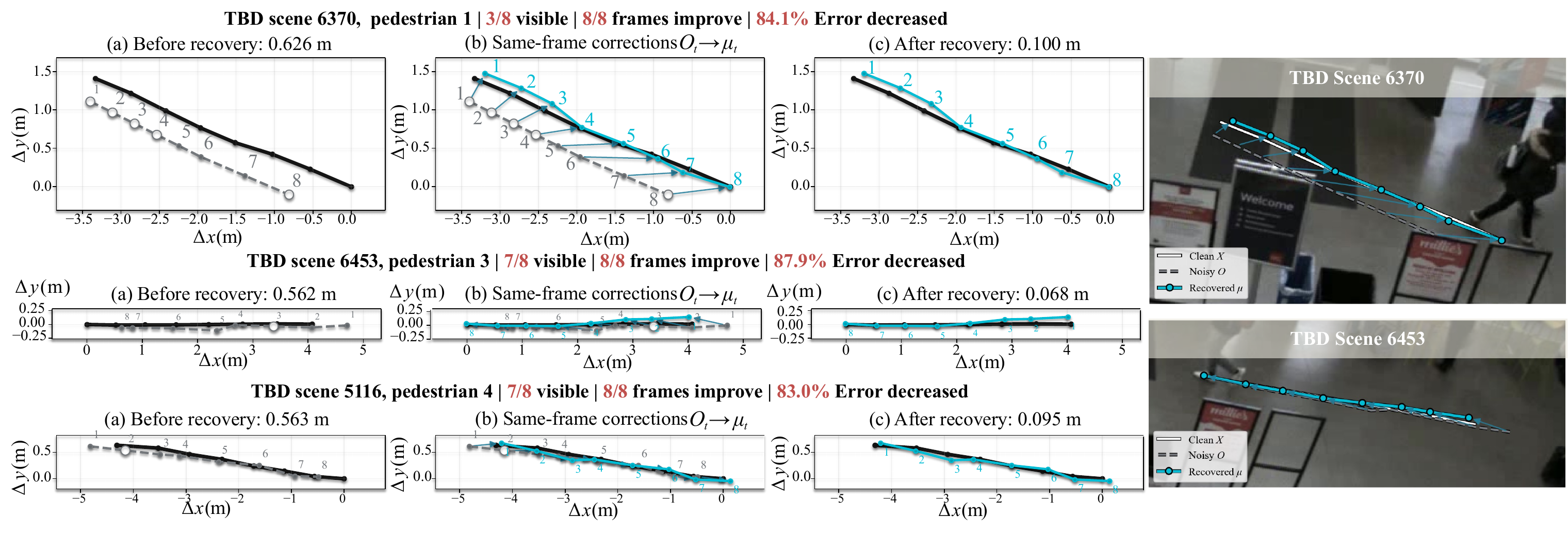}
    \caption{Qualitative history recovery on EgoTraj-TBD. Each row shows (a) noisy observation $O$, (b) framewise corrections $O_t\rightarrow\mu_t$, and (c) recovered history $\mu$ against ground truth $X$. RGB overlays illustrate the first two examples. Hollow markers indicate missing frames and numbers denote time steps.}
    \label{fig:qualitative_tbd_scene}
\end{figure*}

\textbf{Observation-grounded social context.}
Adding social attention to the recovery-and-curriculum variant reduces ADE@1/FDE@1 from $0.967/1.657$ to $0.923/1.593$\,m.
Introducing the residual geometry adapter further reduces these errors to $0.866/1.470$\,m and lowers minADE@20/minFDE@20 from $0.307/0.448$ to $0.289/0.416$\,m.
These results demonstrate the effectiveness of the geometry adapter in integrating observation evidence into the forecasting context.

\textbf{Observation-grounded context across scenes.}
Table~\ref{tab:geometry_final} extends the social-context ablation to the five T2FPV-ETH splits.
Compared with recovered-history conditioning and the curriculum alone, adding the geometry adapter and social attention improves ADE@1, FDE@1, minADE@20, and minFDE@20 on every split.
AVG top-ranked errors decrease from $1.77/2.61$ to $1.52/2.32$\,m, while AVG best-of-20 errors decrease from $0.64/0.81$ to $0.60/0.76$\,m.
These results support the effectiveness of observation-grounded social context across different T2FPV-ETH scenes.

\begin{table}[t]
\centering
\caption{Quantitative results of observation-grounded social context on T2FPV-ETH~\cite{stoler2023t2fpv}. $^{\dagger}$denotes R+C. $^{\S}$ denotes R+C+A+G, as defined in Table~\ref{tab:tbd_core_ablation}.}
\label{tab:geometry_final}
\footnotesize
\setlength{\tabcolsep}{2pt}
\renewcommand{\arraystretch}{1.0} 
\makeatletter
\newcommand{\obrefinegrayrow}[1]{%
  \smash{\rlap{\begingroup%
    \setbox0=\hbox{HOTEL}%
    \setbox2=\hbox{ZARA1}%
    \dimen0=\wd0\relax%
    \ifdim\wd2>\dimen0 \dimen0=\wd2\relax\fi%
    \setbox4=\hbox{#1}%
    \dimen2=\dimexpr\dimen0-\wd4\relax%
    \divide\dimen2 by2\relax%
    \edef\obrefineoffset{\the\dimexpr-\dimen2\relax}%
    \edef\obrefinewidth{\the\columnwidth}%
    \kern\obrefineoffset\relax%
    \color{black!6}%
    \rule[-\dp\@arstrutbox]{\obrefinewidth}%
         {\dimexpr\ht\@arstrutbox+\dp\@arstrutbox\relax}%
  \endgroup}}%
}
\makeatother
\begin{tabular*}{\columnwidth}{@{\extracolsep{\fill}}cccc@{}}
\toprule
\textbf{Split} & \textbf{Variant} & \textbf{@1} & \textbf{@20} \\
\midrule
 & Curriculum$^{\dagger}$ & 1.74/2.82 & 0.72/1.05 \\
\multirow{-2}{*}{ETH} & ObsBridge$^{\S}$ & \textbf{1.47}/\textbf{2.50} & \textbf{0.70}/\textbf{1.01} \\
\addlinespace[1pt]
\obrefinegrayrow{} & Curriculum & 1.33/1.94 & 0.58/0.70 \\
\obrefinegrayrow{HOTEL}\multirow{-2}{*}{HOTEL} & ObsBridge & \textbf{1.29}/\textbf{1.88} & \textbf{0.50}/\textbf{0.60} \\
\addlinespace[1pt]
 & Curriculum & 1.28/2.03 & 0.42/0.55 \\
\multirow{-2}{*}{ZARA1} & ObsBridge & \textbf{1.20}/\textbf{1.96} & \textbf{0.41}/\textbf{0.54} \\
\addlinespace[1pt]
\obrefinegrayrow{} & Curriculum & 1.48/2.10 & 0.53/0.63 \\
\obrefinegrayrow{ZARA2}\multirow{-2}{*}{ZARA2} & ObsBridge & \textbf{1.39}/\textbf{2.06} & \textbf{0.50}/\textbf{0.60} \\
\addlinespace[1pt]
 & Curriculum & 3.04/4.17 & 0.94/1.12 \\
\multirow{-2}{*}{UNIV} & ObsBridge & \textbf{2.26}/\textbf{3.20} & \textbf{0.89}/\textbf{1.07} \\
\midrule
\obrefinegrayrow{} & Curriculum & 1.77/2.61 & 0.64/0.81 \\
\obrefinegrayrow{AVG}\multirow{-2}{*}{AVG} & ObsBridge & \textbf{1.52}/\textbf{2.32} & \textbf{0.60}/\textbf{0.76} \\
\bottomrule
\end{tabular*}
\vspace{-3.6em}
\end{table}

\subsection{Results by Historical Visibility}
\label{sec:exp_visibility}

Table~\ref{tab:visibility_groups} evaluates forecasting accuracy by agent role and pedestrian history visibility on EgoTraj-TBD.
Of the 2,920 pedestrian windows, 2,494 ($85.4\%$) contain incomplete histories.
ObsBridge improves both minADE@20 and minFDE@20 in the 1--3 and 4--7 visible-frame groups.
For pedestrians visible in only 1--3 history frames, errors decrease from $0.69/0.90$ to $0.64/0.80$\,m, corresponding to reductions of $7.2\%/11.1\%$.
Aggregate pedestrian errors decrease from $0.57/0.75$ to $0.54/0.69$\,m, while ego errors decrease from $0.20/0.34$ to $0.18/0.30$\,m.
These results support the effectiveness of ObsBridge under the incomplete historical observations prevalent in this benchmark.

Figure~\ref{fig:qualitative_tbd_scene} further illustrates the effectiveness of historical recovery under noisy and incomplete observations.
History ADE measures the mean Euclidean position error relative to the ground truth over all eight historical time steps, including missing-observation frames.
Across three selected examples, the recovered histories reduce this error by 83.0--87.9\%, with improvements at every time step.
These examples demonstrate that historical recovery can provide more accurate motion histories for subsequent forecasting.

\begin{table}[t]
\centering
\caption{Performance by historical visibility on EgoTraj-TBD~\cite{liu2026egotraj} in minADE@20/minFDE@20 (m, $\downarrow$). Count denotes the number of evaluated agent windows.}
\label{tab:visibility_groups}
\footnotesize
\setlength{\tabcolsep}{2pt}
\renewcommand{\arraystretch}{1.0} 
\makeatletter
\newcommand{\obvisibilitygrayrow}[1]{%
  \smash{\rlap{\begingroup%
    \setbox0=\hbox{\quad 1--3 visible frames}%
    \setbox2=\hbox{\quad 4--7 visible frames}%
    \dimen0=\wd0\relax%
    \ifdim\wd2>\dimen0 \dimen0=\wd2\relax\fi%
    \setbox4=\hbox{#1}%
    \dimen2=\dimexpr\dimen0-\wd4\relax%
    \divide\dimen2 by2\relax%
    \edef\obvisibilityoffset{\the\dimexpr-\dimen2\relax}%
    \edef\obvisibilitywidth{\the\columnwidth}%
    \kern\obvisibilityoffset\relax%
    \color{black!6}%
    \rule[-\dp\@arstrutbox]{\obvisibilitywidth}%
         {\dimexpr\ht\@arstrutbox+\dp\@arstrutbox\relax}%
  \endgroup}}%
}
\makeatother
\begin{tabular*}{\columnwidth}{@{\extracolsep{\fill}}cccc@{}}
\toprule
\textbf{Group} & \textbf{Count} & \textbf{BiFlow}~\cite{liu2026egotraj} & \textbf{ObsBridge} \\
\midrule
Ego & 6,908 & 0.20/0.34 & \textbf{0.18}/\textbf{0.30} \\
\obvisibilitygrayrow{All pedestrians}All pedestrians & 2,920 & 0.57/0.75 & \textbf{0.54}/\textbf{0.69} \\
\midrule
\quad 1--3 visible frames & 969 & 0.69/0.90 & \textbf{0.64}/\textbf{0.80} \\
\obvisibilitygrayrow{\quad 4--7 visible frames}\quad 4--7 visible frames & 1,525 & 0.56/0.73 & \textbf{0.53}/\textbf{0.68} \\
\quad 8 visible frames & 426 & \textbf{0.34}/0.46 & 0.36/\textbf{0.45} \\
\bottomrule
\end{tabular*}
\end{table}

\section{Conclusion}
\label{sec:conclusion}
We presented ObsBridge, which combines recovered-history conditioning, progressive training, and observation-grounded context for ego-view trajectory forecasting.
ObsBridge employs an agent-wise teacher-to-estimate curriculum to adapt the forecaster to imperfect recovered histories, while a residual geometry adapter and social attention integrate geometric, visibility, and correction cues into the forecasting context.
Experiments on EgoTraj-TBD and the five T2FPV-ETH splits demonstrate improvements over the retrained BiFlow baseline in both top-ranked and best-of-20 forecasting.
Specifically, ObsBridge reduces minADE@20/minFDE@20 by 6.5\%/8.7\% on EgoTraj-TBD and by 6.3\%/7.3\% on average across T2FPV-ETH.
Ablation and stratified analyses further support the benefits of observation-grounded conditioning, including for sparsely observed pedestrian histories and multi-agent scenes.
These results support the effectiveness of combining recovered motion with its underlying observation evidence for trajectory forecasting on the ego-view benchmarks.
Future work will investigate generalization across different perception-noise types and real-world deployment environments.

{
    \small
    \bibliographystyle{ieeenat_fullname}
    \bibliography{main}
}

\end{document}